\documentclass[10pt,twocolumn,letterpaper]{article}

\usepackage{cvpr}
\usepackage{times}
\usepackage{epsfig}
\usepackage{graphicx}
\usepackage{amsmath}
\usepackage{amssymb}
\usepackage{tabulary}
\usepackage{bbm}
\usepackage{parskip}

\usepackage[pagebackref=true,breaklinks=true,letterpaper=true,colorlinks,bookmarks=false]{hyperref}

\cvprfinalcopy 

\def\cvprPaperID{1960} 

\ifcvprfinal\fi
\begin{document}

\pagenumbering{gobble}

\title{Deep Visual-Semantic Alignments for Generating Image Descriptions}

\author{
Andrej Karpathy \hspace{0.4in} Li Fei-Fei\\
Department of Computer Science, Stanford University\\
\texttt{\small \{karpathy,feifeili\}@cs.stanford.edu}\\
}

\maketitle

\begin{abstract}
We present a model that generates natural language descriptions of images and their regions. Our approach leverages datasets of images and their sentence descriptions to learn about the inter-modal correspondences between language and visual data. Our alignment model is based on a novel combination of Convolutional Neural Networks over image regions, bidirectional Recurrent Neural Networks over sentences, and a structured objective that aligns the two modalities through a multimodal embedding. We then describe a Multimodal Recurrent Neural Network architecture that uses the inferred alignments to learn to generate novel descriptions of image regions. We demonstrate that our alignment model produces state of the art results in retrieval experiments on Flickr8K, Flickr30K and MSCOCO datasets. We then show that the generated descriptions significantly outperform retrieval baselines on both full images and on a new dataset of region-level annotations.
\end{abstract}


\vspace{-0.2in}
\section{Introduction}
\vspace{-0.1in}

A quick glance at an image is sufficient for a human to point out and describe an immense amount of details about the visual scene \cite{fei2007we}. However, this remarkable ability has proven to be an elusive task for our visual recognition models. The majority of previous work in visual recognition has focused on labeling images with a fixed set of visual categories and great progress has been achieved in these endeavors \cite{ilsvrc,Everingham10}. However, while closed vocabularies of visual concepts constitute a convenient modeling assumption, they are vastly restrictive when compared to the enormous amount of rich descriptions that a human can compose.

Some pioneering approaches that address the challenge of generating image descriptions have been developed \cite{kulkarni2011baby,farhadi2010every}. However, these models often rely on hard-coded visual concepts and sentence templates, which imposes limits on their variety. Moreover, the focus of these works has been on reducing complex visual scenes into a single sentence, which we consider to be an unnecessary restriction.

\begin{figure}[t]
\includegraphics[width=1\linewidth]{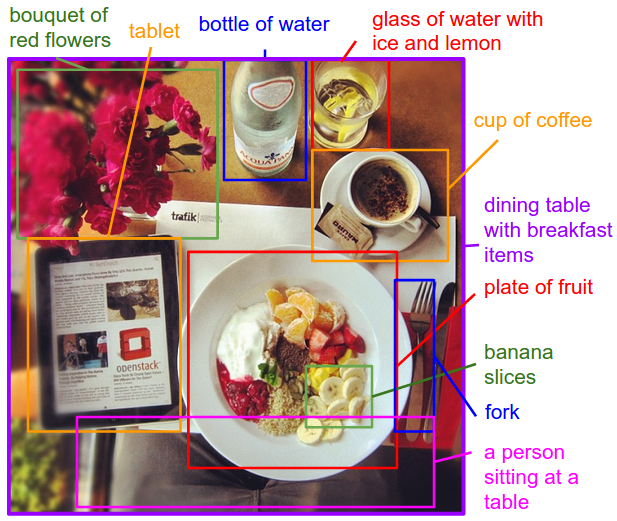}
\caption{Motivation/Concept Figure: Our model treats language as a rich label space and generates descriptions of image regions.}
\label{fig:pull}
\vspace{-0.1in}
\end{figure}

In this work, we strive to take a step towards the goal of generating dense descriptions of images (Figure \ref{fig:pull}). The primary challenge towards this goal is in the design of a model that is rich enough to simultaneously reason about contents of images and their representation in the domain of natural language. Additionally, the model should be free of assumptions about specific hard-coded templates, rules or categories and instead rely on learning from the training data. The second, practical challenge is that datasets of image captions are available in large quantities on the internet \cite{hodosh2013framing,flickr30k,coco}, but these descriptions multiplex mentions of several entities whose locations in the images are unknown.

Our core insight is that we can leverage these large image-sentence datasets by treating the sentences as weak labels, in which contiguous segments of words correspond to some particular, but unknown location in the image. Our approach is to infer these alignments and use them to learn a generative model of descriptions. Concretely, our contributions are twofold:

\begin{itemize}
\item We develop a deep neural network model that infers the latent alignment between segments of sentences and the region of the image that they describe. Our model associates the two modalities through a common, multimodal embedding space and a structured objective. We validate the effectiveness of this approach on image-sentence retrieval experiments in which we surpass the state-of-the-art.
\item We introduce a multimodal Recurrent Neural Network architecture that takes an input image and generates its description in text. Our experiments show that the generated sentences significantly outperform retrieval-based baselines, and produce sensible qualitative predictions. We then train the model on the inferred correspondences and evaluate its performance on a new dataset of region-level annotations.
\end{itemize}

We make code, data and annotations publicly available. \footnote{\texttt{cs.stanford.edu/people/karpathy/deepimagesent}}

\vspace{-0.1in}
\section{Related Work}
\vspace{-0.1in}

\textbf{Dense image annotations.} Our work shares the high-level goal of densely annotating the contents of images with many works before us. Barnard et al. \cite{barnard2003matching} and Socher et al. \cite{socher2010connecting} studied the multimodal correspondence between words and images to annotate segments of images. Several works \cite{li2009towards,gould2009decomposing,fidler2013sentence,li2007and} studied the problem of holistic scene understanding in which the scene type, objects and their spatial support in the image is inferred. However, the focus of these works is on correctly labeling scenes, objects and regions with a fixed set of categories, while our focus is on richer and higher-level descriptions of regions.

\textbf{Generating descriptions.} The task of describing images with sentences has also been explored. A number of approaches pose the task as a retrieval problem, where the most compatible annotation in the training set is transferred to a test image \cite{hodosh2013framing,sochergrounded,farhadi2010every,ordonez2011im2text,JiaICCV11}, or where training annotations are broken up and stitched together \cite{Kuznetsova2012,lisiming2011,kuznetsova2014treetalk}. Several approaches generate image captions based on fixed templates that are filled based on the content of the image \cite{gupta2012image,kulkarni2011baby,farhadi2010every,yang2011corpus,yao2010i2t,elliott2013image,barbu2012video} or generative grammars  \cite{Mitchell2012,yatskar2014see}, but this approach limits the variety of possible outputs. Most closely related to us, Kiros et al. \cite{kirosmultimodal} developed a log-bilinear model that can generate full sentence descriptions for images, but their model uses a fixed window context while our Recurrent Neural Network (RNN) model conditions the probability distribution over the next word in a sentence on all previously generated words. Multiple closely related preprints appeared on Arxiv during the submission of this work, some of which also use RNNs to generate image descriptions \cite{mao2014explain,vinyals2014show,donahue2014long,kiros2014unifying,fang2014captions,chen14}. Our RNN is simpler than most of these approaches but also suffers in performance. We quantify this comparison in our experiments.

\textbf{Grounding natural language in images.} A number of approaches have been developed for grounding text in the visual domain \cite{kong2014you,matuszek2012,zitnicklearning,linvisual}. Our approach is inspired by Frome et al. \cite{frome2013devise} who associate words and images through a semantic embedding. More closely related is the work of Karpathy et al. \cite{defrag}, who decompose images and sentences into fragments and infer their inter-modal alignment using a ranking objective. In contrast to their model which is based on grounding dependency tree relations, our model aligns contiguous segments of sentences which are more meaningful, interpretable, and not fixed in length.

\textbf{Neural networks in visual and language domains.} Multiple approaches have been developed for representing images and words in higher-level representations. On the image side, Convolutional Neural Networks (CNNs) \cite{lecun1998gradient,krizhevsky2012imagenet} have recently emerged as a powerful class of models for image classification and object detection \cite{ilsvrc}. On the sentence side, our work takes advantage of pretrained word vectors \cite{mikolov2013distributed,jeffreypenningtonglove,bengio2006neural} to obtain low-dimensional representations of words. Finally, Recurrent Neural Networks have been previously used in language modeling \cite{mikolov2010recurrent,sutskever2011generating}, but we additionally condition these models on images.

\vspace{-0.1in}
\section{Our Model}
\vspace{-0.15in}

\begin{figure*}[t]
\includegraphics[width=1\linewidth]{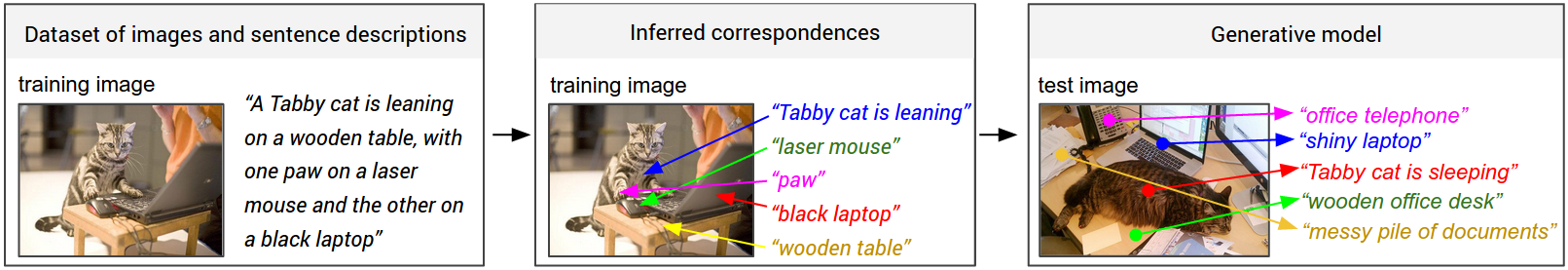}
\caption{Overview of our approach. A dataset of images and their sentence descriptions is the input to our model (left). Our model first infers the correspondences (middle, Section \ref{sec:align}) and then learns to generate novel descriptions (right, Section \ref{sec:generate}).}
\vspace{-0.1in}
\label{fig:sys1}
\end{figure*}

\textbf{Overview}. The ultimate goal of our model is to generate descriptions of image regions. During training, the input to our model is a set of images and their corresponding sentence descriptions (Figure \ref{fig:sys1}). We first present a model that aligns sentence snippets to the visual regions that they describe through a multimodal embedding. We then treat these correspondences as training data for a second, multimodal Recurrent Neural Network model that learns to generate the snippets.

\vspace{-0.1in}
\subsection{Learning to align visual and language data}
\label{sec:align}
\vspace{-0.15in}

Our alignment model assumes an input dataset of images and their sentence descriptions. Our key insight is that sentences written by people make frequent references to some particular, but unknown location in the image. For example, in Figure \ref{fig:sys1}, the words \textit{``Tabby cat is leaning''} refer to the cat, the words \textit{``wooden table''} refer to the table, etc. We would like to infer these latent correspondences, with the eventual goal of later learning to generate these snippets from image regions. We build on the approach of Karpathy et al. \cite{defrag}, who learn to ground dependency tree relations to image regions with a ranking objective. Our contribution is in the use of bidirectional recurrent neural network to compute word representations in the sentence, dispensing of the need to compute dependency trees and allowing unbounded interactions of words and their context in the sentence. We also substantially simplify their objective and show that both modifications improve ranking performance.

We first describe neural networks that map words and image regions into a common, multimodal embedding. Then we introduce our novel objective, which learns the embedding representations so that semantically similar concepts across the two modalities occupy nearby regions of the space.

\vspace{-0.15in}
\subsubsection{Representing images}
\vspace{-0.15in}

Following prior work \cite{kulkarni2011baby,defrag}, we observe that sentence descriptions make frequent references to objects and their attributes. Thus, we follow the method of Girshick et al. \cite{girshick2014rcnn} to detect objects in every image with a Region Convolutional Neural Network (RCNN). The CNN is pre-trained on ImageNet \cite{deng2009imagenet} and finetuned on the 200 classes of the ImageNet Detection Challenge \cite{ilsvrc}. Following Karpathy et al. \cite{defrag}, we use the top 19 detected locations in addition to the whole image and compute the representations based on the pixels $I_b$ inside each bounding box as follows:

\vspace{-0.15in}
\begin{equation}
v = W_m [\text{{\it CNN}}_{\theta_c}(I_b)] + b_m,
\end{equation}
\vspace{-0.2in}

where $\text{{\it CNN}}(I_b)$ transforms the pixels inside bounding box $I_b$ into 4096-dimensional activations of the fully connected layer immediately before the classifier. The CNN parameters $\theta_c$ contain approximately 60 million parameters. The matrix $W_m$ has dimensions $h \times 4096$, where $h$ is the size of the multimodal embedding space ($h$ ranges from 1000-1600 in our experiments). Every image is thus represented as a set of $h$-dimensional vectors $\{v_i \mid i = 1 \ldots 20\}$.

\vspace{-0.1in}
\subsubsection{Representing sentences}
\vspace{-0.15in}

To establish the inter-modal relationships, we would like to represent the words in the sentence in the same $h$-dimensional embedding space that the image regions occupy. The simplest approach might be to project every individual word directly into this embedding. However, this approach does not consider any ordering and word context information in the sentence. An extension to this idea is to use word bigrams, or dependency tree relations as previously proposed \cite{defrag}. However, this still imposes an arbitrary maximum size of the context window and requires the use of Dependency Tree Parsers that might be trained on unrelated text corpora.

To address these concerns, we propose to use a Bidirectional Recurrent Neural Network (BRNN) \cite{schuster1997bidirectional} to compute the word representations. The BRNN takes a sequence of $N$ words (encoded in a 1-of-k representation) and transforms each one into an $h$-dimensional vector. However, the representation of each word is enriched by a variably-sized context around that word. Using the index $t = 1 \ldots N$ to denote the position of a word in a sentence, the precise form of the BRNN is as follows:

\vspace{-0.22in}
\begin{align}
& x_t = W_w \mathbbm{I}_t \\
& e_t = f(W_e x_t + b_e) \\
& h_t^f = f(e_t + W_f h^f_{t-1} + b_f) \\
& h_t^b = f(e_t + W_b h^b_{t+1} + b_b) \\
& s_t = f(W_d ( h_t^f + h_t^b ) + b_d).
\end{align}
\vspace{-0.25in}

\begin{figure}
\centering
\includegraphics[width=0.75\linewidth]{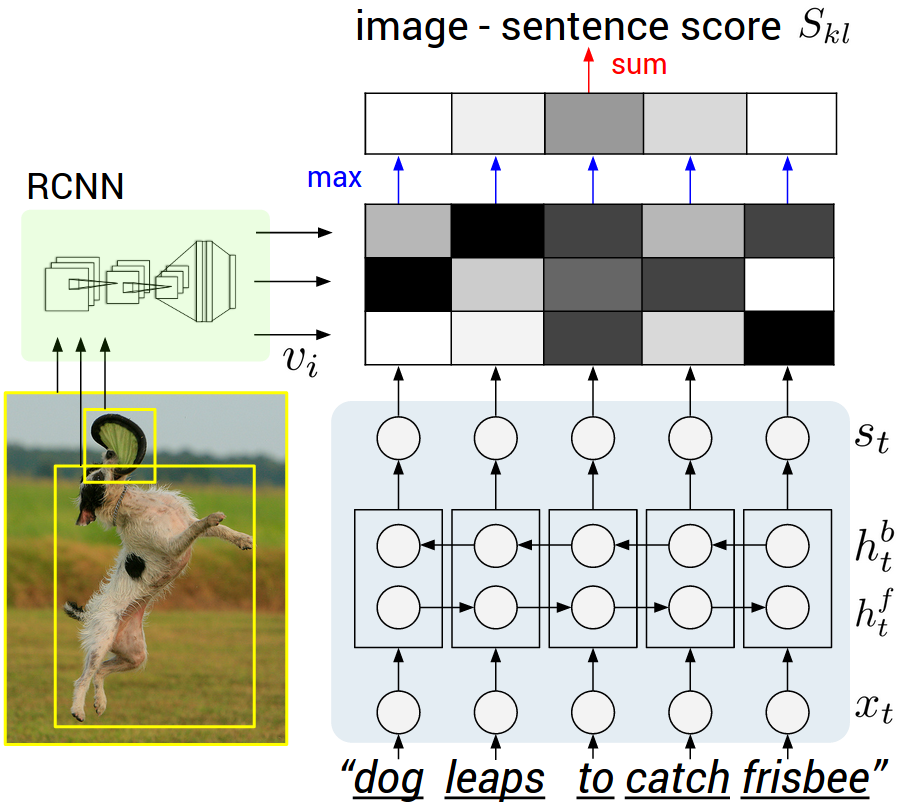}
\caption{Diagram for evaluating the image-sentence score $S_{kl}$. Object regions are embedded with a CNN (left). Words (enriched by their context) are embedded in the same multimodal space with a BRNN (right). Pairwise similarities are computed with inner products (magnitudes shown in grayscale) and finally reduced to image-sentence score with Equation \ref{eq:skl}.}
\label{fig:brnn}
\vspace{-0.1in}
\end{figure}

Here, $\mathbbm{I}_t$ is an indicator column vector that has a single one at the index of the $t$-th word in a word vocabulary. The weights $W_w$ specify a word embedding matrix that we initialize with 300-dimensional word2vec \cite{mikolov2013distributed} weights and keep fixed due to overfitting concerns. However, in practice we find little change in final performance when these vectors are trained, even from random initialization. Note that the BRNN consists of two independent streams of processing, one moving left to right ($h_t^f$) and the other right to left ($h_t^b$) (see Figure \ref{fig:brnn} for diagram). The final $h$-dimensional representation $s_t$ for the $t$-th word is a function of both the word at that location and also its surrounding context in the sentence. Technically, every $s_t$ is a function of all words in the entire sentence, but our empirical finding is that the final word representations ($s_t$) align most strongly to the visual concept of the word at that location ($\mathbbm{I}_t$).

We learn the parameters $W_e, W_f, W_b, W_d$ and the respective biases $b_e, b_f, b_b, b_d$. A typical size of the hidden representation in our experiments ranges between 300-600 dimensions. We set the activation function $f$ to the rectified linear unit (ReLU), which computes $f: x \mapsto max(0, x)$.

\vspace{-0.1in}
\subsubsection{Alignment objective}
\label{sec:rankingloss}
\vspace{-0.1in}

We have described the transformations that map every image and sentence into a set of vectors in a common $h$-dimensional space. Since the supervision is at the level of entire images and sentences, our strategy is to formulate an image-sentence score as a function of the individual region-word scores. Intuitively, a sentence-image pair should have a high matching score if its words have a confident support in the image. The model of Karpathy et a. \cite{defrag} interprets the dot product $v_i^Ts_t$ between the $i$-th region and $t$-th word as a measure of similarity and use it to define the score between image $k$ and sentence $l$ as:

\vspace{-0.15in}
\begin{equation}
S_{kl} = \sum_{t \in g_l}\sum_{i \in g_k} max(0, v_i^Ts_t).
\end{equation}
\vspace{-0.15in}

Here, $g_k$ is the set of image fragments in image $k$ and $g_l$ is the set of sentence fragments in sentence $l$. The indices $k,l$ range over the images and sentences in the training set. Together with their additional Multiple Instance Learning objective, this score carries the interpretation that a sentence fragment aligns to a subset of the image regions whenever the dot product is positive. We found that the following reformulation simplifies the model and alleviates the need for additional objectives and their hyperparameters:

\vspace{-0.15in}
\begin{equation}
S_{kl} = \sum_{t \in g_l}max_{i \in g_k} v_i^Ts_t.
\label{eq:skl}
\end{equation}
\vspace{-0.15in}

Here, every word $s_t$ aligns to the single best image region. As we show in the experiments, this simplified model also leads to improvements in the final ranking performance. Assuming that $k=l$ denotes a corresponding image and sentence pair, the final max-margin, structured loss remains:

\vspace{-0.15in}
\begin{align}
\mathcal{C}(\theta) = \sum_k & \Big[ \underbrace{ \sum_l max(0, S_{kl} - S_{kk} + 1) }_\text{rank images} \\
& + \underbrace { \sum_l max(0, S_{lk} - S_{kk} + 1) }_\text{rank sentences} \Big] \notag.
\end{align}
\vspace{-0.15in}

This objective encourages aligned image-sentences pairs to have a higher score than misaligned pairs, by a margin.

\vspace{-0.1in}
\subsubsection{Decoding text segment alignments to images}
\label{sec:mrf}
\vspace{-0.1in}

Consider an image from the training set and its corresponding sentence. We can interpret the quantity $v_i^Ts_t$ as the unnormalized log probability of the $t$-th word describing any of the bounding boxes in the image. However, since we are ultimately interested in generating snippets of text instead of single words, we would like to align extended, contiguous sequences of words to a single bounding box. Note that the na\"{i}ve solution that assigns each word independently to the highest-scoring region is insufficient because it leads to words getting scattered inconsistently to different regions.

To address this issue, we treat the true alignments as latent variables in a Markov Random Field (MRF) where the binary interactions between neighboring words encourage an alignment to the same region. Concretely, given a sentence with $N$ words and an image with $M$ bounding boxes, we introduce the latent alignment variables $a_j \in {\{1 \ldots M\}}$ for $j=1\ldots N$ and formulate an MRF in a chain structure along the sentence as follows:

\vspace{-0.2in}
\begin{align}
& E(\textbf{a}) = \sum_{j=1 \ldots N} \psi_j^U(a_j) + \sum_{j=1 \ldots N-1} \psi_{j}^B(a_j, a_{j+1}) \\
& \psi_j^U(a_j = t) = v_i^Ts_t \\
& \psi_{j}^B(a_j, a_{j+1}) = \beta \mathbbm{1}{ [a_j = a_{j+1}] }.
\end{align}
\vspace{-0.2in}

Here, $\beta$ is a hyperparameter that controls the affinity towards longer word phrases. This parameter allows us to interpolate between single-word alignments ($\beta = 0$) and aligning the entire sentence to a single, maximally scoring region when $\beta$ is large. We minimize the energy to find the best alignments $\textbf{a}$ using dynamic programming. The output of this process is a set of image regions annotated with segments of text. We now describe an approach for generating novel phrases based on these correspondences.

\vspace{-0.1in}
\subsection{Multimodal Recurrent Neural Network for generating descriptions}
\label{sec:generate}
\vspace{-0.1in}

In this section we assume an input set of images and their textual descriptions. These could be full images and their sentence descriptions, or regions and text snippets, as inferred in the previous section. The key challenge is in the design of a model that can predict a variable-sized sequence of outputs given an image. In previously developed language models based on Recurrent Neural Networks (RNNs) \cite{mikolov2010recurrent,sutskever2011generating,elman1990finding}, this is achieved by defining a probability distribution of the next word in a sequence given the current word and context from previous time steps. We explore a simple but effective extension that additionally conditions the generative process on the content of an input image. More formally, during training our Multimodal RNN takes the image pixels $I$ and a sequence of input vectors $(x_1, \ldots, x_T)$. It then computes a sequence of hidden states $(h_1, \ldots , h_t)$ and a sequence of outputs $(y_1, \dots , y_t)$ by iterating the following recurrence relation for $t = 1$ to $T$:

\vspace{-0.3in}
\begin{align}
\label{eq:rnn}
& b_v = W_{hi} [\text{{\it CNN}}_{\theta_c}(I)] \\
& h_t = f(W_{hx} x_{t} + W_{hh} h_{t-1} + b_h + \mathbbm{1}{(t=1)} \odot b_v) \\
& y_t =softmax( W_{oh} h_t + b_o).
\end{align}
\vspace{-0.3in}

In the equations above, $W_{hi}, W_{hx}, W_{hh}, W_{oh}, x_i$ and $b_h, b_o$ are learnable parameters, and $\text{{\it CNN}}_{\theta_c}(I)$ is the last layer of a CNN. The output vector $y_t$ holds the (unnormalized) log probabilities of words in the dictionary and one additional dimension for a special END token. Note that we provide the image context vector $b_v$ to the RNN only at the first iteration, which we found to work better than at each time step. In practice we also found that it can help to also pass both $b_v, (W_{hx} x_{t})$ through the activation function. A typical size of the hidden layer of the RNN is 512 neurons.

\begin{figure}
\centering
\includegraphics[width=0.6\linewidth]{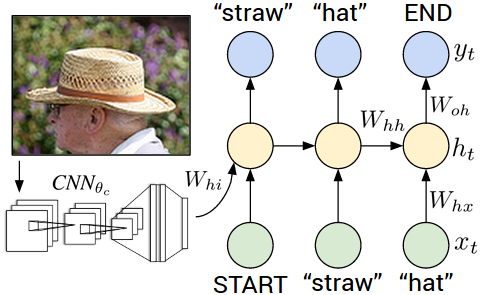}
\caption{Diagram of our multimodal Recurrent Neural Network generative model. The RNN takes a word, the context from previous time steps and defines a distribution over the next word in the sentence. The RNN is conditioned on the image information at the first time step. START and END are special tokens.}
\vspace{-0.1in}
\label{fig:rnn}
\end{figure}

\textbf{RNN training.} The RNN is trained to combine a word ($x_t$), the previous context ($h_{t-1}$)  to predict the next word ($y_t$). We condition the RNN's predictions on the image information ($b_v$) via bias interactions on the first step. The training proceeds as follows (refer to Figure \ref{fig:rnn}): We set $h_0 = \vec{0}$, $x_1$ to a special START vector, and the desired label $y_1$ as the first word in the sequence. Analogously, we set $x_2$ to the word vector of the first word and expect the network to predict the second word, etc. Finally, on the last step when $x_T$ represents the last word, the target label is set to a special END token. The cost function is to maximize the log probability assigned to the target labels (i.e. Softmax classifier).

\textbf{RNN at test time.} To predict a sentence, we compute the image representation $b_v$, set $h_0 = 0$, $x_1$ to the START vector and compute the distribution over the first word $y_1$. We sample a word from the distribution (or pick the argmax), set its embedding vector as $x_2$, and repeat this process until the END token is generated. In practice we found that beam search (e.g. beam size 7) can improve results.

\vspace{-0.05in}
\subsection{Optimization}
\vspace{-0.15in}

We use SGD with mini-batches of 100 image-sentence pairs and momentum of 0.9 to optimize the alignment model. We cross-validate the learning rate and the weight decay. We also use dropout regularization in all layers except in the recurrent layers \cite{zaremba2014recurrent} and clip gradients elementwise at 5 (important). The generative RNN is more difficult to optimize, party due to the word frequency disparity between rare words and common words (e.g. "a" or the END token). We achieved the best results using RMSprop \cite{rmsprop}, which is an adaptive step size method that scales the update of each weight by a running average of its gradient norm.

\begin{table*}[t]
\small
\centering
\begin{tabulary}{\linewidth}{L|CCCC|CCCC}
& \multicolumn{4}{c}{Image Annotation} & \multicolumn{4}{c}{Image Search} \\
\textbf{Model} & \textbf{R@1} & \textbf{R@5} & \textbf{R@10} & \textbf{Med} \it{r} & \textbf{R@1} & \textbf{R@5} & \textbf{R@10} & \textbf{Med} \it{r} \\
\hline
\multicolumn{9}{c}{\textbf{Flickr30K}} \\
\hline
SDT-RNN (Socher et al. \cite{sochergrounded}) & 9.6 & 29.8 & 41.1 & 16 & 8.9 & 29.8 & 41.1 & 16 \\
Kiros et al. \cite{kiros2014unifying} & 14.8 & 39.2 & 50.9 & 10 & 11.8 & 34.0 & 46.3 & 13 \\
Mao et al. \cite{mao2014explain} & 18.4 & 40.2 & 50.9 & 10 & 12.6 & 31.2 & 41.5 & 16 \\
Donahue et al. \cite{donahue2014long} & 17.5 & 40.3 & 50.8 & 9 & - & - & - & - \\
DeFrag (Karpathy et al. \cite{defrag}) & 14.2 & 37.7 & 51.3 & 10 & 10.2 & 30.8 & 44.2 & 14 \\
Our implementation of DeFrag \cite{defrag} & 19.2 & 44.5 & 58.0 & 6.0 & 12.9 & 35.4 & 47.5 & 10.8\\
Our model: DepTree edges & 20.0 & 46.6 & 59.4 & 5.4 & 15.0 & 36.5 & 48.2 & 10.4\\
Our model: BRNN & \textbf{22.2} & \textbf{48.2} & \textbf{61.4} & \textbf{4.8} & \textbf{15.2} & \textbf{37.7} & \textbf{50.5} & \textbf{9.2} \\
\hline
Vinyals et al. \cite{vinyals2014show} (more powerful CNN) & 23 & - & 63 & 5 & 17 & - & 57 & 8 \\
\hline
\multicolumn{9}{c}{\textbf{MSCOCO}} \\
\hline
Our model: 1K test images & 38.4 & 69.9 & 80.5 & 1.0 & 27.4 & 60.2 & 74.8 & 3.0 \\
Our model: 5K test images & 16.5 & 39.2 & 52.0 & 9.0 & 10.7 & 29.6 & 42.2 & 14.0 \\
\hline
\end{tabulary}
\vspace{0.05in}
\caption{{\small Image-Sentence ranking experiment results. \textbf{R@K} is Recall@K (high is good). \textbf{Med} {\it r} is the median rank (low is good). In the results for our models, we take the top 5 validation set models, evaluate each independently on the test set and then report the average performance. The standard deviations on the recall values range from approximately 0.5 to 1.0.}}
\label{tab:ranking}
\vspace{-0.1in}
\end{table*}

\vspace{-0.1in}
\section{Experiments}
\vspace{-0.15in}

\textbf{Datasets.} We use the Flickr8K \cite{hodosh2013framing}, Flickr30K \cite{flickr30k} and MSCOCO \cite{coco} datasets in our experiments. These datasets contain 8,000, 31,000 and 123,000 images respectively and each is annotated with 5 sentences using Amazon Mechanical Turk. For Flickr8K and Flickr30K, we use 1,000 images for validation, 1,000 for testing and the rest for training (consistent with \cite{hodosh2013framing,defrag}). For MSCOCO we use 5,000 images for both validation and testing. \\

\vspace{-0.2in}
\textbf{Data Preprocessing.} We convert all sentences to lowercase, discard non-alphanumeric characters. We filter words to those that occur at least 5 times in the training set, which results in 2538, 7414, and 8791 words for Flickr8k, Flickr30K, and MSCOCO datasets respectively.

\vspace{-0.1in}
\subsection{Image-Sentence Alignment Evaluation}
\vspace{-0.15in}
\label{sec:ranking}

\begin{figure*}[t]
\includegraphics[width=1\linewidth]{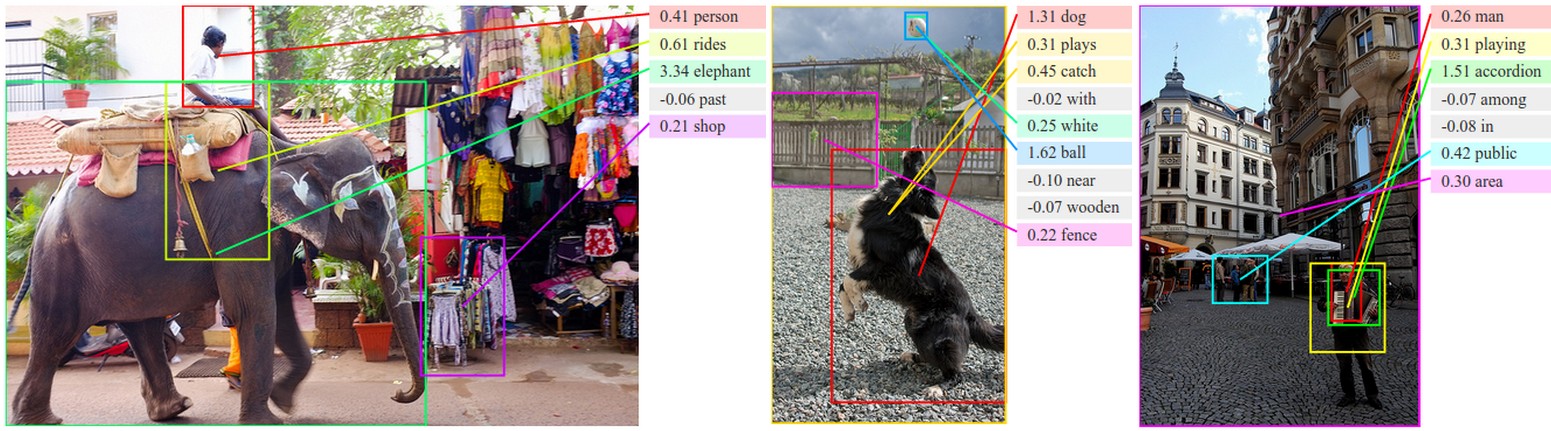}
\caption{Example alignments predicted by our model. For every test image above, we retrieve the most compatible test sentence and visualize the highest-scoring region for each word (before MRF smoothing described in Section \ref{sec:mrf}) and the associated scores ($v_i^Ts_t$). We hide the alignments of low-scoring words to reduce clutter. We assign each region an arbitrary color.}
\vspace{-0.1in}
\label{fig:ranking}
\end{figure*}

We first investigate the quality of the inferred text and image alignments with ranking experiments. We consider a withheld set of images and sentences and retrieve items in one modality given a query from the other by sorting based on the image-sentence score $S_{kl}$ (Section \ref{sec:rankingloss}). We report the median rank of the closest ground truth result in the list and Recall@K, which measures the fraction of times a correct item was found among the top K results. The result of these experiments can be found in Table \ref{tab:ranking}, and example retrievals in Figure \ref{fig:ranking}. We now highlight some of the takeaways.

\textbf{Our full model outperforms previous work.} First, our full model (``Our model: BRNN'') outperforms Socher et al. \cite{sochergrounded} who trained with a similar loss but used a single image representation and a Recursive Neural Network over the sentence. A similar loss was adopted by Kiros et al. \cite{kiros2014unifying}, who use an LSTM \cite{hochreiter1997long} to encode sentences. We list their performance with a CNN that is equivalent in power (AlexNet \cite{krizhevsky2012imagenet}) to the one used in this work, though similar to \cite{vinyals2014show} they outperform our model with a more powerful CNN (VGGNet \cite{simonyan2014very}, GoogLeNet \cite{szegedy2014going}). ``DeFrag'' are the results reported by Karpathy et al. \cite{defrag}. Since we use different word vectors, dropout for regularization and different cross-validation ranges and larger embedding sizes, we re-implemented their loss for a fair comparison (``Our implementation of DeFrag''). Compared to other work that uses AlexNets, our full model shows consistent improvement.

\vspace{-0.05in}
\textbf{Our simpler cost function improves performance.} We strive to better understand the source of our performance. First, we removed the BRNN and used dependency tree relations exactly as described in Karpathy et al. \cite{defrag} (``Our model: DepTree edges''). The only difference between this model and ``Our reimplementation of DeFrag'' is the new, simpler cost function introduced in Section \ref{sec:rankingloss}. We see that our formulation shows consistent improvements.

\vspace{-0.05in}
\textbf{BRNN outperforms dependency tree relations}. Furthermore, when we replace the dependency tree relations with the BRNN we observe additional performance improvements. Since the dependency relations were shown to work better than single words and bigrams \cite{defrag}, this suggests that the BRNN is taking advantage of contexts longer than two words. Furthermore, our method does not rely on extracting a Dependency Tree and instead uses the raw words directly.

\vspace{-0.05in}
\textbf{MSCOCO results for future comparisons.} We are not aware of other published ranking results on MSCOCO. Therefore, we report results on a subset of 1,000 images and the full set of 5,000 test images for future comparisons. Note that the 5000 images numbers are lower since Recall@K is a function of test set size.

\vspace{-0.05in}
\textbf{Qualitative.} As can be seen from example groundings in Figure \ref{fig:ranking}, the model discovers interpretable visual-semantic correspondences, even for small or relatively rare objects such as an \textit{``accordion''}. These would be likely missed by models that only reason about full images.

\vspace{-0.05in}
\textbf{Learned region and word vector magnitudes.} An appealing feature of our model is that it learns to modulate the magnitude of the region and word embeddings. Due to their inner product interaction, we observe that representations of visually discriminative words such as \textit{``kayaking, pumpkins``} have embedding vectors with higher magnitudes, which in turn  translates to a higher influence on the image-sentence score. Conversely, stop words such as \textit{``now, simply, actually, but''} are mapped near the origin, which reduces their influence. See more analysis in supplementary material.

\begin{table*}
\small
\centering
\begin{tabulary}{\linewidth}{L|CCCC|CCCC|CCCCCC}
\hline
& \multicolumn{4}{c}{\textbf{Flickr8K}} & \multicolumn{4}{c}{\textbf{Flickr30K}} & \multicolumn{6}{c}{\textbf{MSCOCO 2014}}\\
\hline 
\textbf{Model} & B-1 & B-2 & B-3 & B-4 & B-1 & B-2 & B-3 & B-4 & B-1 & B-2 & B-3& B-4 & METEOR & CIDEr \\
\hline
Nearest Neighbor & --- & --- & --- & --- & --- & --- & --- & --- & 48.0 & 28.1 & 16.6 & 10.0 & 15.7 & 38.3\\
Mao et al. \cite{mao2014explain} &58 & 28 & 23 & --- & 55 & 24 & 20 & --- & --- & --- & --- &  --- & --- & --- \\
Google NIC \cite{vinyals2014show} & 63 & 41 & 27& --- & 66.3 & 42.3 & 27.7& 18.3 & 66.6 & 46.1 & 32.9 & 24.6 & --- & --- \\
LRCN \cite{donahue2014long} & --- & --- & --- & --- & 58.8 & 39.1 & 25.1 & 16.5 & 62.8 & 44.2 & 30.4 & --- & --- & --- \\
MS Research \cite{fang2014captions} & --- & --- & --- & --- & --- & --- & --- & --- & --- & --- & --- & 21.1 & 20.7& --- \\
Chen and Zitnick \cite{chen14} & --- & --- & --- & 14.1 & --- & --- & --- & 12.6 & --- & --- & --- & 19.0 &20.4 & --- \\
Our model & 57.9 & 38.3 & 24.5  & 16.0 & 57.3  & 36.9  & 24.0  &15.7 & 62.5 & 45.0 & 32.1 & 23.0 & 19.5 & 66.0 \\
\hline
\end{tabulary}
\vspace{0.01in}
\caption{Evaluation of full image predictions on 1,000 test images. \textbf{B-n} is BLEU score that uses up to n-grams. High is good in all columns. For future comparisons, our METEOR/CIDEr Flickr8K scores are 16.7/31.8 and the Flickr30K scores are 15.3/24.7.}
\label{tab:bleu}
\vspace{-0.1in}
\end{table*}

\begin{figure*}[t]
\includegraphics[width=\linewidth]{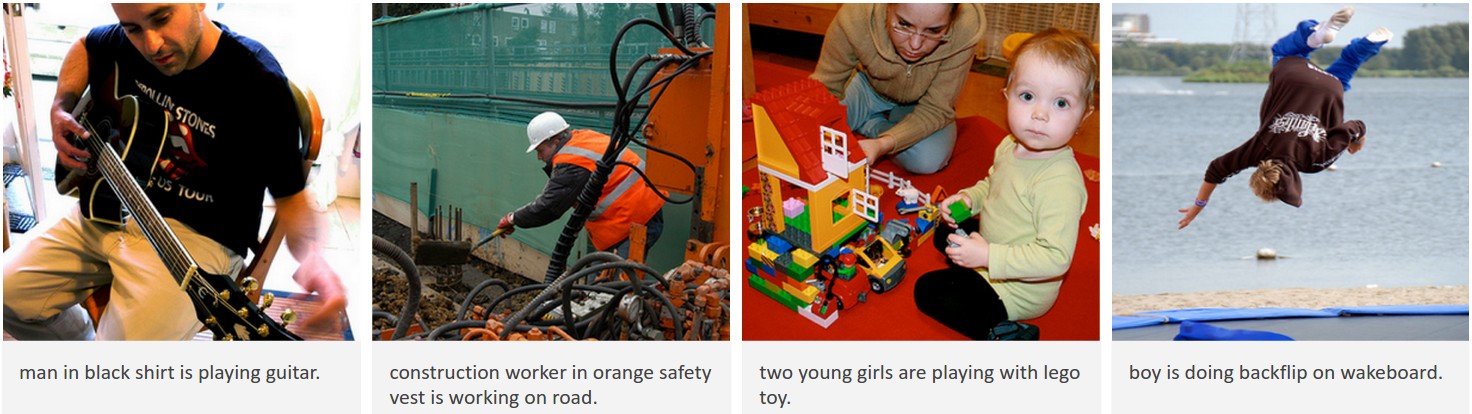}
\caption{Example sentences generated by the multimodal RNN for test images. We provide many more examples on our project page.}
\label{fig:gen}
\vspace{-0.1in}
\end{figure*}

\subsection{Generated Descriptions: Fulframe evaluation}
\vspace{-0.15in}

We now evaluate the ability of our RNN model to describe images and regions. We first trained our Multimodal RNN to generate sentences on full images with the goal of verifying that the model is rich enough to support the mapping from image data to sequences of words. For these full image experiments we use the more powerful VGGNet image features \cite{simonyan2014very}.  We report the BLEU \cite{papineni2002bleu}, METEOR \cite{meteor} and CIDEr \cite{cider} scores computed with the \texttt{coco-caption} code \cite{capeval2015} \footnote{\texttt{https://github.com/tylin/coco-caption}}. Each method evaluates a \textit{candidate} sentence by measuring how well it matches a set of five \textit{reference} sentences written by humans. 

\begin{figure*}[t]
\includegraphics[width=\linewidth]{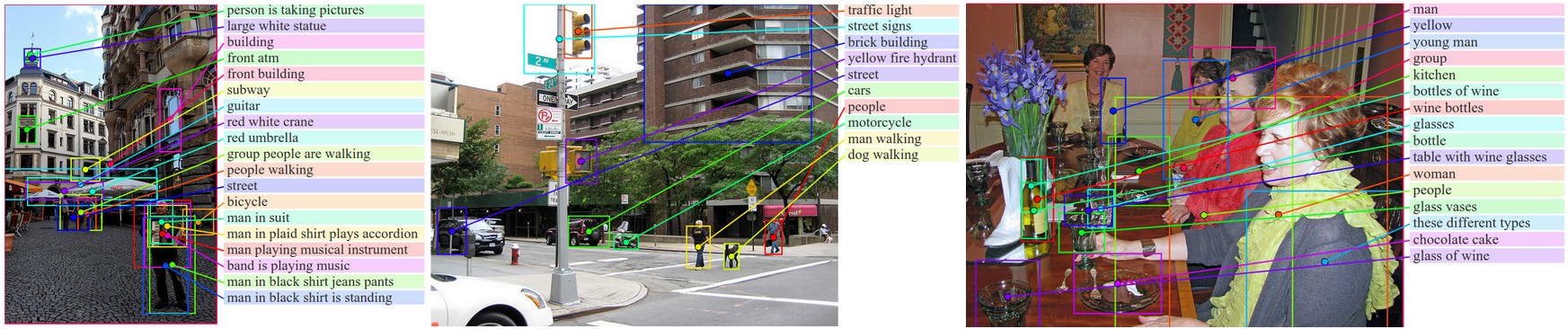}
\caption{Example region predictions. We use our region-level multimodal RNN to generate text (shown on the right of each image) for some of the bounding boxes in each image. The lines are grounded to centers of bounding boxes and the colors are chosen arbitrarily.}
\label{fig:genregion}
\vspace{-0.15in}
\end{figure*}

\textbf{Qualitative.} The model generates sensible descriptions of images (see Figure \ref{fig:gen}), although we consider the last two images failure cases. The first prediction \textit{``man in black shirt is playing a guitar''} does not appear in the training set. However, there are 20 occurrences of ``man in black shirt'' and 60 occurrences of ``is paying guitar'', which the model may have composed to describe the first image. In general, we find that a relatively large portion of generated sentences (60\% with beam size 7) can be found in the training data. This fraction decreases with lower beam size; For instance, with beam size 1 this falls to 25\%, but the performance also deteriorates (e.g. from 0.66 to 0.61 CIDEr).

\textbf{Multimodal RNN outperforms retrieval baseline.} Our first comparison is to a nearest neighbor retrieval baseline. Here, we annotate each test image with a sentence of the most similar training set image as determined by L2 norm over VGGNet \cite{simonyan2014very} fc7 features. Table \ref{tab:bleu} shows that the Multimodal RNN confidently outperforms this retrieval method. Hence, even with 113,000 train set images in MSCOCO the retrieval approach is inadequate. Additionally, the RNN takes only a fraction of a second to evaluate per image.

\textbf{Comparison to other work.} Several related models have been proposed in Arxiv preprints since the original submission of this work. We also include these in Table \ref{tab:bleu} for comparison. Most similar to our model is Vinyals et al. \cite{vinyals2014show}. Unlike this work where the image information is communicated through a bias term on the first step, they incorporate it as a first word, they use a more powerful but more complex sequence learner (LSTM \cite{hochreiter1997long}), a different CNN (GoogLeNet \cite{szegedy2014going}), and report results of a model ensemble.  Donahue et al. \cite{donahue2014long} use a 2-layer factored LSTM (similar in structure to the RNN in Mao et al. \cite{mao2014explain}). Both models appear to work worse than ours, but this is likely in large part due to their use of the less powerful AlexNet \cite{krizhevsky2012imagenet} features. Compared to these approaches, our model prioritizes simplicity and speed at a slight cost in performance.

\vspace{-0.1in}
\subsection{Generated Descriptions: Region evaluation}
\vspace{-0.15in}

We now train the Multimodal RNN on the correspondences between image regions and snippets of text, as inferred by the alignment model. To support the evaluation, we used Amazon Mechanical Turk (AMT) to collect a new dataset of region-level annotations that we only use at test time. The labeling interface displayed a single image and asked annotators (we used nine per image) to draw five bounding boxes and annotate each with text. In total, we collected 9,000 text snippets for 200 images in our MSCOCO test split (i.e. 45 snippets per image). The snippets have an average length of 2.3 words. Example annotations include \textit{``sports car'', ``elderly couple sitting'', ``construction site'', ``three dogs on leashes'', ``chocolate cake''}. We noticed that asking annotators for grounded text snippets induces language statistics different from those in full image captions. Our region annotations are more comprehensive and feature elements of scenes that would rarely be considered salient enough to be included in a single sentence sentence about the full image, such as \textit{``heating vent'', ``belt buckle'', and ``chimney''}.

\textbf{Qualitative}. We show example region model predictions in Figure \ref{fig:genregion}. To reiterate the difficulty of the task, consider for example the phrase \textit{``table with wine glasses''} that is generated on the image on the right in Figure \ref{fig:genregion}. This phrase only occurs in the training set 30 times. Each time it may have a different appearance and each time it may occupy a few (or none) of our object bounding boxes. To generate this string for the region, the model had to first correctly learn to ground the string and then also learn to generate it.

\textbf{Region model outperforms full frame model and ranking baseline}. Similar to the full image description task, we evaluate this data as a prediction task from a 2D array of pixels (one image region) to a sequence of words and record the BLEU score. The ranking baseline retrieves training sentence substrings most compatible with each region as judged by the BRNN model. Table \ref{tab:bleu2} shows that the region RNN model produces descriptions most consistent with our collected data. Note that the fullframe model was trained only on full images, so feeding it smaller image regions deteriorates its performance. However, its sentences are also longer than the region model sentences, which likely negatively impacts the BLEU score. The sentence length is non-trivial to control for with an RNN, but we note that the region model also outperforms the fullframe model on all other metrics: CIDEr 61.6/20.3, METEOR 15.8/13.3, ROUGE 35.1/21.0 for region/fullframe respectively.

\begin{table}
\small
\centering
\begin{tabulary}{\linewidth}{L|CCCC}
\hline
\textbf{Model} & B-1 & B-2 & B-3 & B-4\\
\hline
Human agreement & 61.5 & 45.2 & 30.1& 22.0 \\
\hline
Nearest Neighbor & 22.9 & 10.5 & 0.0 & 0.0 \\
RNN: Fullframe model & 14.2 & 6.0 & 2.2& 0.0 \\
RNN: Region level model & \textbf{35.2} & \textbf{23.0} & \textbf{16.1} & \textbf{14.8} \\
\hline
\end{tabulary}
\vspace{0.05in}
\caption{BLEU score evaluation of image region annotations.}
\label{tab:bleu2}
\vspace{-0.2in}
\end{table}

\vspace{-0.1in}
\subsection{Limitations}
\vspace{-0.15in}

Although our results are encouraging, the Multimodal RNN model is subject to multiple limitations. First, the model can only generate a description of one input array of pixels at a fixed resolution. A more sensible approach might be to use multiple saccades around the image to identify all entities, their mutual interactions and wider context before generating a description. Additionally, the RNN receives the image information only through additive bias interactions, which are known to be less expressive than more complicated multiplicative interactions \cite{sutskever2011generating,hochreiter1997long}. Lastly, our approach consists of two separate models. Going directly from an image-sentence dataset to region-level annotations as part of a single model trained end-to-end remains an open problem.

\vspace{-0.15in}
\section{Conclusions}
\vspace{-0.15in}

We introduced a model that generates natural language descriptions of image regions based on weak labels in form of a dataset of images and sentences, and with very few hard-coded assumptions. Our approach features a novel ranking model that aligned parts of visual and language modalities through a common, multimodal embedding. We showed that this model provides state of the art performance on image-sentence ranking experiments. Second, we described a Multimodal Recurrent Neural Network architecture that generates descriptions of visual data. We evaluated its performance on both fullframe and region-level experiments and showed that in both cases the Multimodal RNN outperforms retrieval baselines.

\textbf{Acknowledgements.}\\
We thank Justin Johnson and Jon Krause for helpful comments and discussions. We gratefully acknowledge the support of NVIDIA Corporation with the donation of the GPUs used for this research. This research is partially supported by an ONR MURI grant, and NSF ISS-1115313.

{\small
\bibliographystyle{ieee}
\bibliography{egbib}
}

\clearpage

\section{Supplementary Material}
\subsection{Magnitude modulation}
\vspace{-0.1in}

An appealing feature of our alignment model is that it learns to modulate the importance of words and regions by scaling the magnitude of their corresponding embedding vectors. To see this, recall that we compute the image-sentence similarity between image $k$ and sentence $l$ as follows:

\vspace{-0.1in}
\begin{equation}
S_{kl} = \sum_{t \in g_l}max_{i \in g_k} v_i^Ts_t.
\end{equation}
\vspace{-0.15in}

\textbf{Disciminative words.} As a result of this formulation, we observe that representations of visually discriminative words such as \textit{``kayaking, pumpkins``} tend to have higher magnitude in the embedding space, which translates to a higher influence on the final image-sentence scores due to the inner product. Conversely, the model learns to map stop words such as \textit{``now, simply, actually, but''} near the origin, which reduces their influence. Table \ref{tab:magnitudes} show the top 40 words with highest and lowest magnitudes $\|s_t\|$.

\begin{figure}[b]
\includegraphics[width=\linewidth]{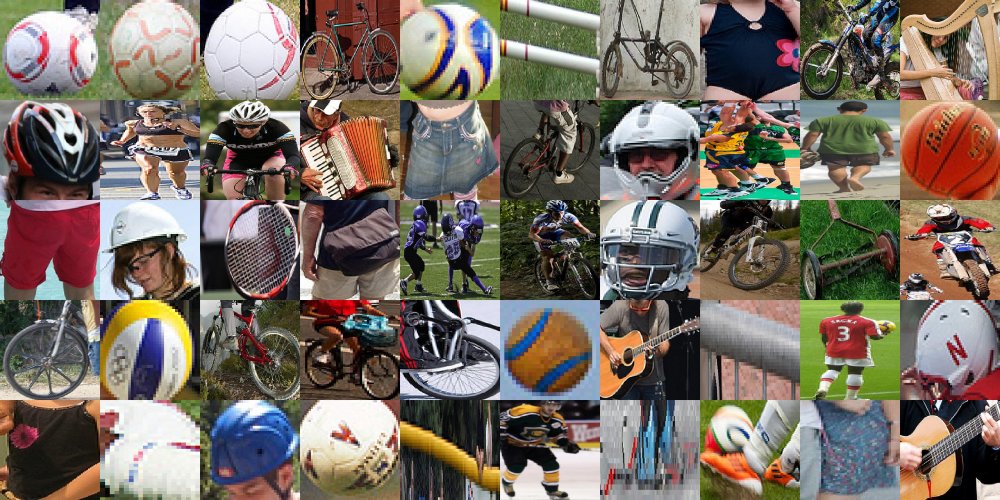}
\caption{Flickr30K test set regions with high vector magnitude.}
\label{fig:outconf}
\end{figure}

\textbf{Disciminative regions.} Similarly, image regions that contain discriminative entities are assigned vectors of higher magnitudes by our model. This can be be interpreted as a measure of visual saliency, since these regions would produced large scores if their textual description was present in a corresponding sentence. We show the regions with high magnitudes in Figure \ref{fig:outconf}. Notice the common occurrence of often described regions such as balls, bikes, helmets.

\begin{table}
\small
\centering
\begin{tabulary}{\linewidth}{R|L|R|L}
\hline
Magnitude & Word & Magnitude & Word \\
\hline
0.42 & now & 2.61 & kayaking \\
0.42 & simply & 2.59 & trampoline \\
0.43 & actually & 2.59 & pumpkins \\
0.44 & but & 2.58 & windsurfing \\
0.44 & neither & 2.56 & wakeboard \\
0.45 & then & 2.54 & acrobatics \\
0.45 & still & 2.54 & sousaphone \\
0.46 & obviously & 2.54 & skydivers \\
0.47 & that & 2.52 & wakeboarders \\
0.47 & which & 2.52 & skateboard \\
0.47 & felt & 2.51 & snowboarder \\
0.47 & not & 2.51 & wakeboarder \\
0.47 & might & 2.50 & skydiving \\
0.47 & because & 2.50 & guitar \\
0.48 & appeared & 2.50 & snowboard \\
0.48 & therefore & 2.48 & kitchen \\
0.48 & been & 2.48 & paraglider \\
0.48 & if & 2.48 & ollie \\
0.48 & also & 2.47 & firetruck \\
0.48 & only & 2.47 & gymnastics \\
0.48 & so & 2.46 & waterfalls \\
0.49 & would & 2.46 & motorboat \\
0.49 & yet & 2.46 & fryer \\
0.50 & be & 2.46 & skateboarding \\
0.50 & had & 2.46 & dulcimer \\
0.50 & revealed & 2.46 & waterfall \\
0.50 & never & 2.46 & backflips \\
0.50 & very & 2.46 & unicyclist \\
0.50 & without & 2.45 & kayak \\
0.51 & they & 2.43 & costumes \\
0.51 & either & 2.43 & wakeboarding \\
0.51 & could & 2.43 & trike \\
0.51 & feel & 2.42 & dancers \\
0.51 & otherwise & 2.42 & cupcakes \\
0.51 & when & 2.42 & tuba \\
0.51 & already & 2.42 & skijoring \\
0.51 & being & 2.41 & firewood \\
0.51 & else & 2.41 & elevators \\
0.52 & just & 2.40 & cranes \\
0.52 & ones & 2.40 & bassoon \\
\hline
\end{tabulary}
\caption{This table shows the top magnitudes of vectors ($\|s_t\|$) for words in Flickr30K. Since the magnitude of individual words in our model is also a function of their surrounding context in the sentence, we report the average magnitude.}
\label{tab:magnitudes}
\end{table}

\clearpage

\subsection{Alignment model}

\begin{figure*}
\large
``chocolate cake''\\
\includegraphics[width=\linewidth]{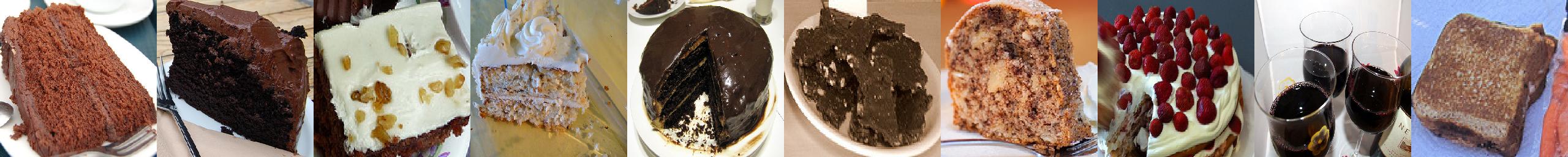}

``glass of wine''\\
\includegraphics[width=\linewidth]{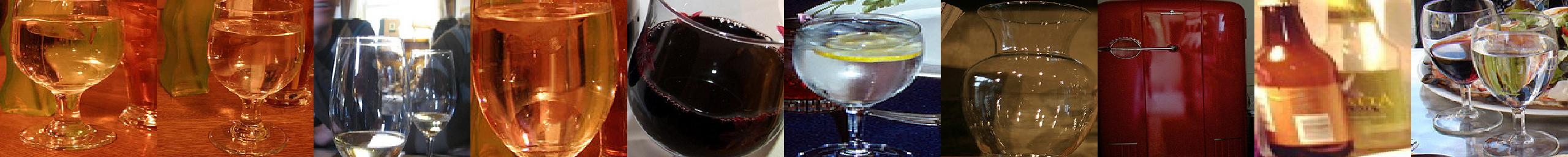}

``red bus''\\
\includegraphics[width=\linewidth]{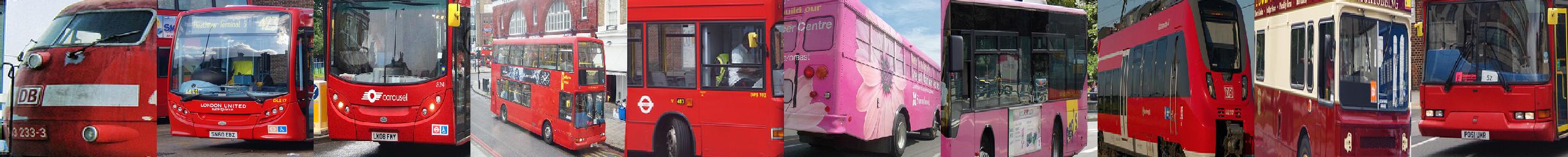}

``yellow bus''\\
\includegraphics[width=\linewidth]{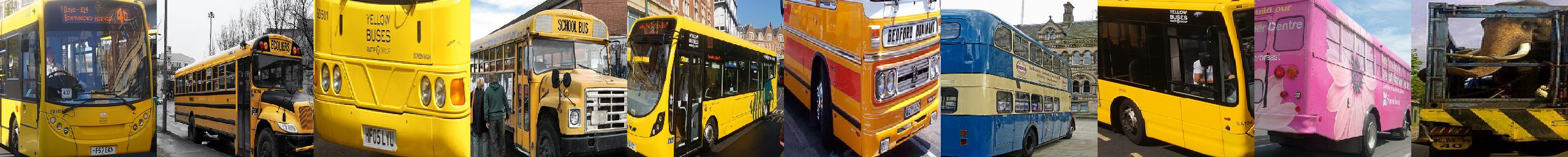}

``closeup of zebra''\\
\includegraphics[width=\linewidth]{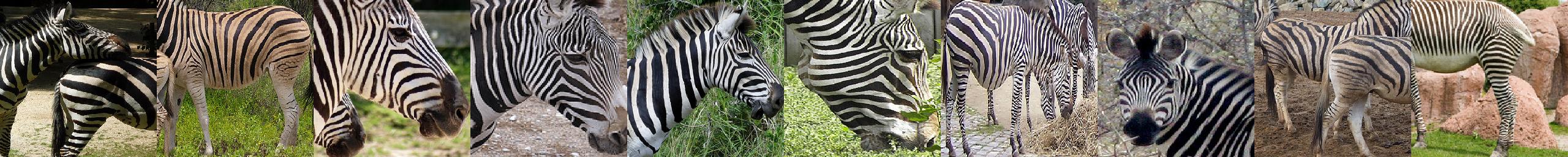}

``sprinkled donut''\\
\includegraphics[width=\linewidth]{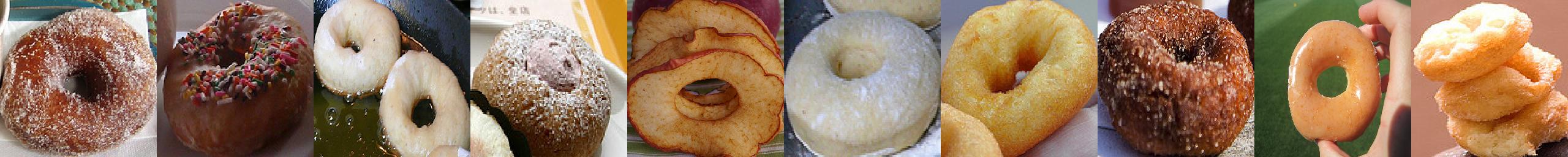}

``wooden chair''\\
\includegraphics[width=\linewidth]{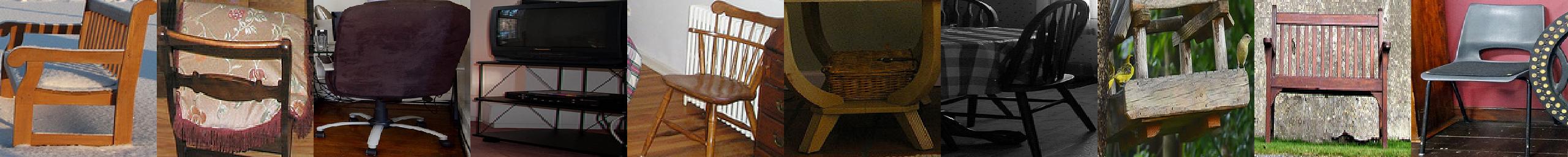}

``wooden office desk''\\
\includegraphics[width=\linewidth]{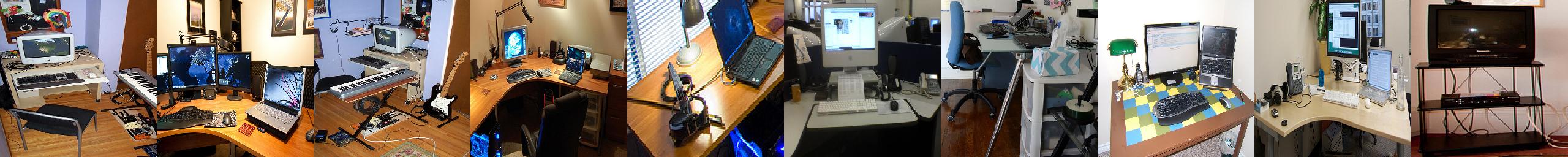}

``shiny laptop''\\
\includegraphics[width=\linewidth]{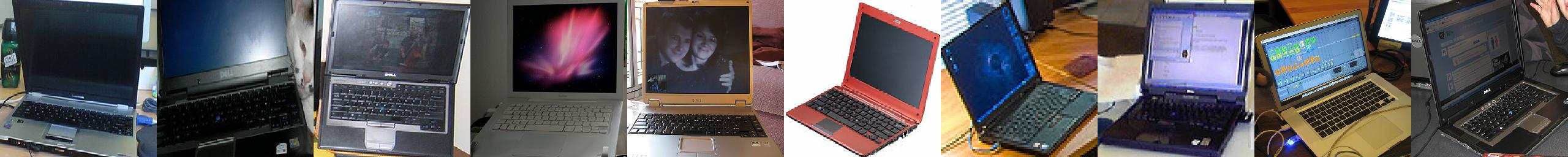}

\caption{Examples of highest scoring regions for queried snippets of text, on 5,000 images of our MSCOCO test set.}
\label{fig:textqueries}
\end{figure*}

\begin{figure*}
\large
``bird flying in the sky''\\
\includegraphics[width=\linewidth]{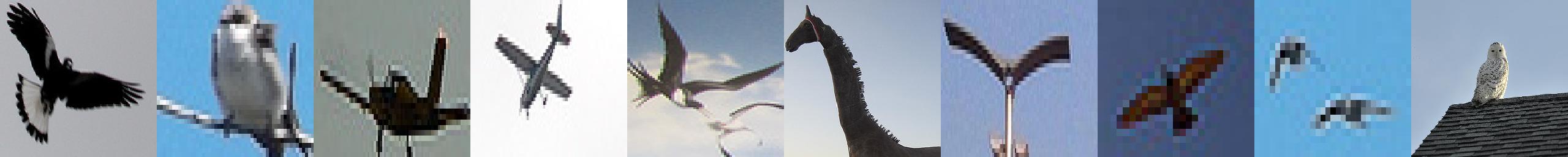}

``bird on a tree branch''\\
\includegraphics[width=\linewidth]{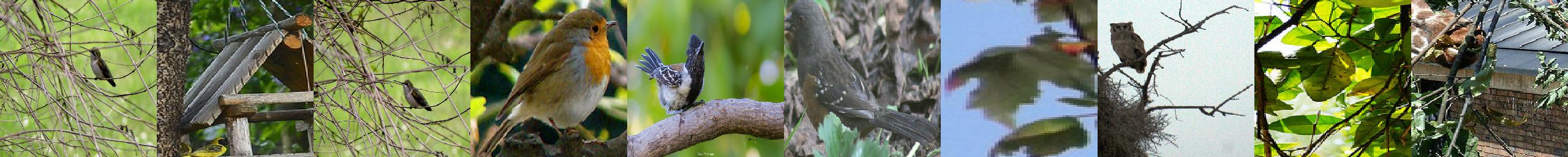}

``bird sitting on roof''\\
\includegraphics[width=\linewidth]{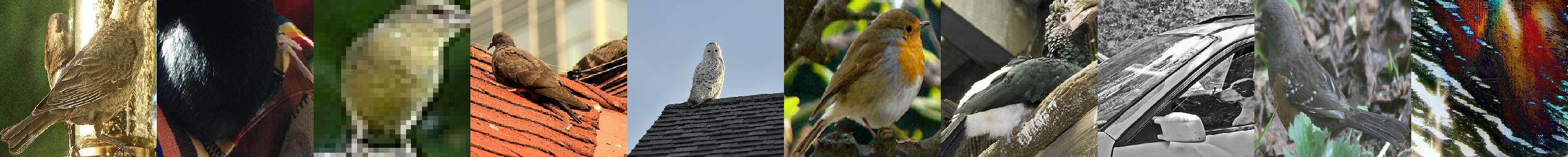}

``closeup of fruit''\\
\includegraphics[width=\linewidth]{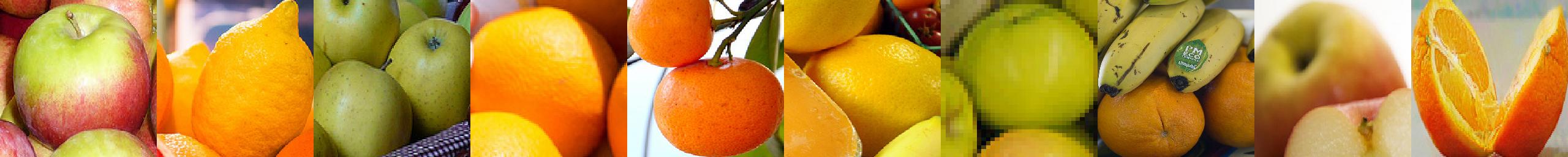}

``bowl of fruit''\\
\includegraphics[width=\linewidth]{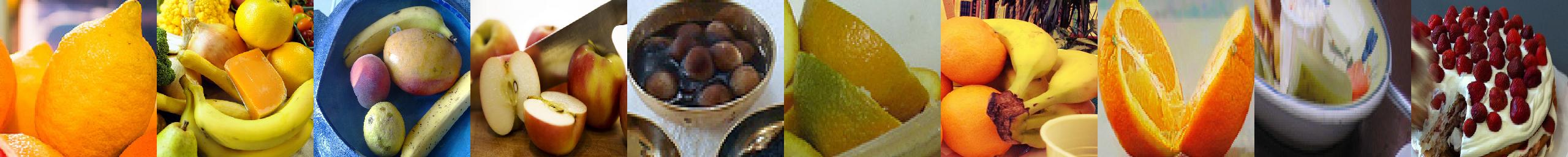}

``man riding a horse''\\
\includegraphics[width=\linewidth]{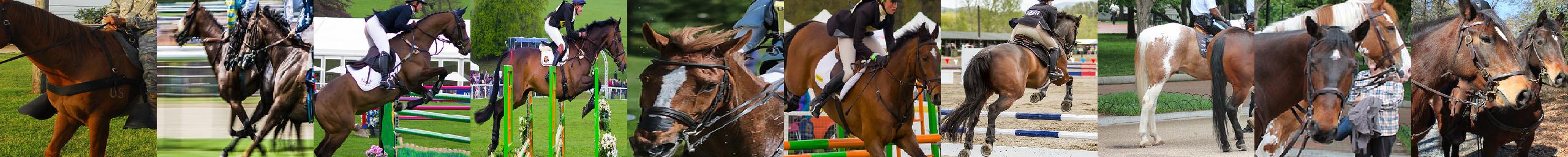}

``straw hat''\\
\includegraphics[width=\linewidth]{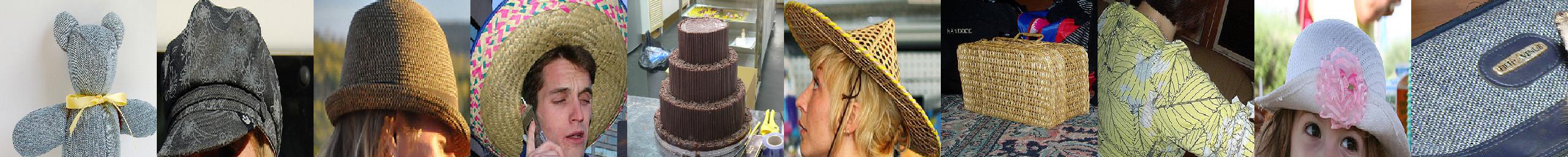}

\caption{Examples of highest scoring regions for queried snippets of text, on 5,000 images of our MSCOCO test set.}
\label{fig:textqueries2}
\end{figure*}

\textbf{Learned appearance of text snippets}. We can query our alignment model with a piece of text and retrieve individual image regions that have the highest score with that snippet. We show examples of such queries in Figure \ref{fig:textqueries} and Figure \ref{fig:textqueries2}. Notice that the model is sensitive to compound words and modifiers. For example, \textit{``red bus''} and \textit{``yellow bus''} give very different results. Similarly, \textit{``bird flying in the sky''} and \textit{``bird on a tree branch''} give different results. Additionally, it can be seen that the quality of the results deteriorates for less frequently occurring concepts, such as \textit{``roof''} or \textit{``straw hat''}. However, we emphasize that the model learned these visual appearances of text snippets from raw data of full images and sentences, without any explicit correspondences.

\begin{figure*}
\includegraphics[width=\linewidth]{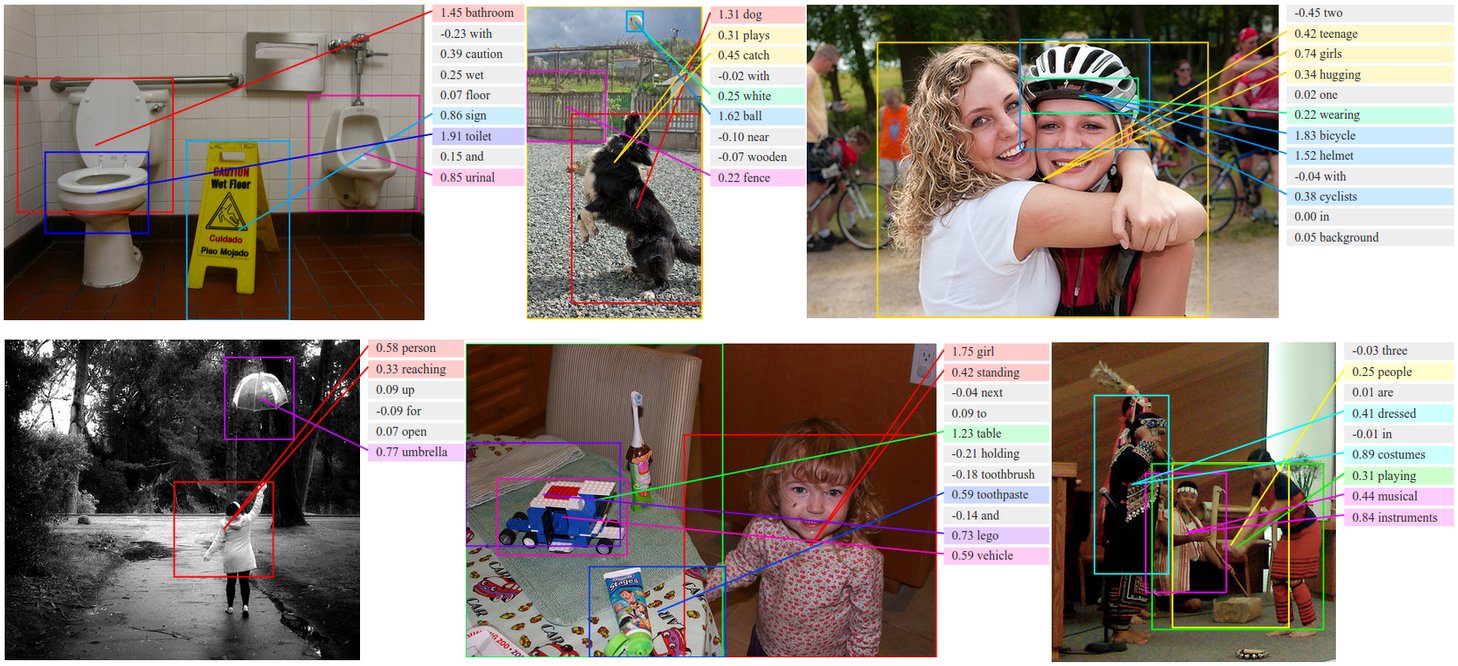}
\caption{Additional examples of alignments. For each query test image above we retrieve the most compatible sentence from the test set and show the alignments.}
\label{fig:alignments2}
\end{figure*}

\textbf{Additional alignment visualizations}. See additional examples of inferred alignments between image regions and words in Figure \ref{fig:alignments2}. Note that one limitation of our model is that it does not explicitly handle or support counting. For instance, the last example we show contains the phrase \textit{``three people''}. These words should align to the three people in the image, but our model puts the bounding box around two of the people. In doing so, the model may be taking advantage of the BRNN structure to modify the ``people'' vector to preferentially align to regions that contain multiple people. However, this is still unsatisfying because such spurious detections only exist as a result of an error in the RCNN inference process, which presumably failed to localize the individual people.

\textbf{Web demo}. We have published a web demo that displays our alignments for all images in the test set \footnote{\tiny \texttt{http://cs.stanford.edu/people/karpathy/deepimagesent/rankingdemo/}}.

\textbf{Additional Flickr8K experiments}. We omitted ranking experiment results from our paper due to space constraints, but these can be found in Table \ref{tab:8k}

\textbf{Counting}. We experimented with losses that perform probabilistic inference in the forward pass that explicitly tried to localize exactly three distinct people in the image. However, this worked poorly because while the RCNN is good at finding people, it is not very good at localizing them. For instance, a single person can easily yield multiple detections (the head, the torso, or the full body, for example). We were not able to come up with a simple approach to collapsing these into a single detection (non-maxim suppression by itself was not sufficient in our experiments). Note that this ambiguity is partly an artifact of the training data. For example, torsos of people can often be labeled alone if the body is occluded. We are therefore lead to believe that this additional modeling step is highly non-trivial and a worthy subject of future work.

\textbf{Plug and play use of Natural Language Processing toolkits.} Before adopting the BRNN approach, we also tried to use Natural Language Processing toolkits to process the input sentences into graphs of noun phrases and their binary relations. For instance, in the sentence \textit{``a brown dog is chasing a young child''}, the toolkit would infer that there are two noun phrases (\textit{``a brown dog'', ``young child''}), joined by a binary relationship of \textit{``chasing''}. We then developed a CRF that inferred the grounding of these noun phrases to the detection bounding boxes in the image with a unary appearance model and a spatial binary model. However, this endeavor proved fruitless. First, performing CRF-like inference during the forward pass of a Neural Network proved to be extremely slow. Second, we found that there is surprisingly little information in the relative spatial positions between bounding boxes. For instance, almost any two bounding boxes in the image could correspond to the action of \textit{``chasing''} due to huge amount of possibly camera views of a scene. Hence, we were unable to extract enough signal from the binary relations in the coordinate system of the image and suspect that more complex 3-dimensional reasoning may be required. Lastly, we found that NLP tools (when used out of the box) introduce a large amount of mistakes in the extracted parse trees, dependency trees and parts of speech tags. We tried to fix these with complex rules and exceptions, but ultimately decided to abandon the idea. We believe that part of the problem is that these tools are usually trained on different text corpora (e.g. news articles), so image captions are outside of their domain of competence. In our experience, adopting the BRNN model instead of this approach provided immediate performance improvements and produced significant reductions in code complexity.

\subsection{Additional examples: Image annotation}
\begin{figure*}
\includegraphics[width=\linewidth]{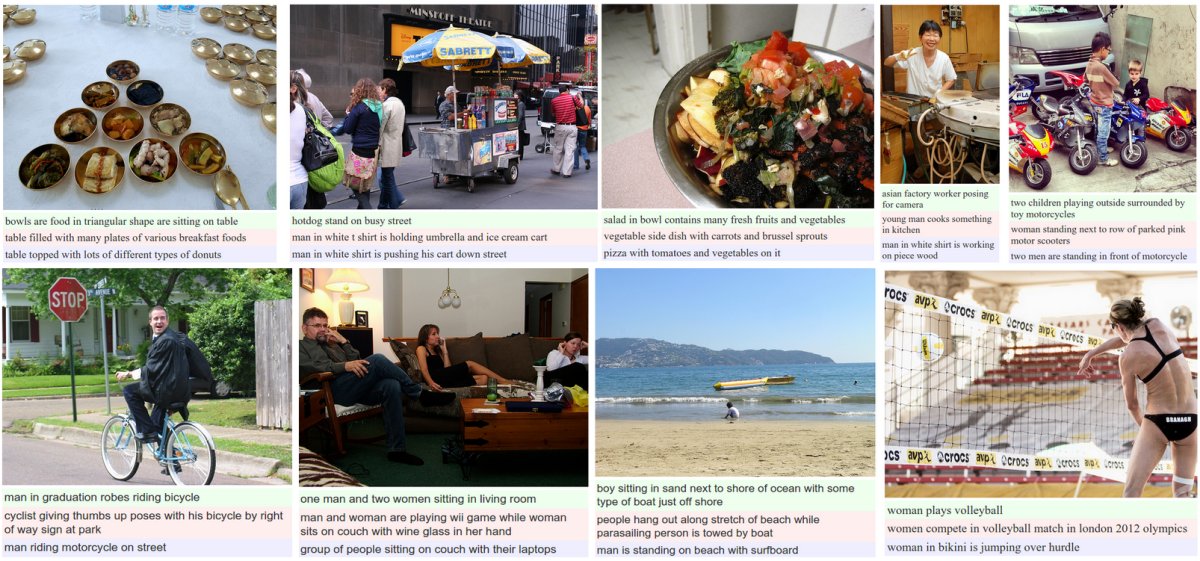}
\caption{Additional examples of captions on the level of full images. Green: Human ground truth. Red: Top-scoring sentence from training set. Blue: Generated sentence.}
\label{fig:fulimage2}
\end{figure*}

Additional examples of generated captions on the full image level can be found in Figure \ref{fig:fulimage2} (and our website). The model often gets the right gist of the scene, but sometimes guesses specific fine-grained words incorrectly. We expect that reasoning not only the global level of the image but also on the level of objects will significantly improve these results. We find the last example (\textit{``woman in bikini is jumping over hurdle''}) to be especially illuminating. This sentence does not occur in the training data. Our general qualitative impression of the model is that it learns certain templates, e.g. \textit{``\textless noun\textgreater in \textless noun\textgreater is \textless verb\textgreater in \textless noun\textgreater''}, and then fills these in based on textures in the image. In this particular case, the volleyball net has the visual appearance of a hurdle, which may have caused the model to insert it as a noun (along with the woman) into one of its learned sentence templates.

\subsection{Additional examples: Region annotation}

\begin{figure*}
\includegraphics[width=\linewidth]{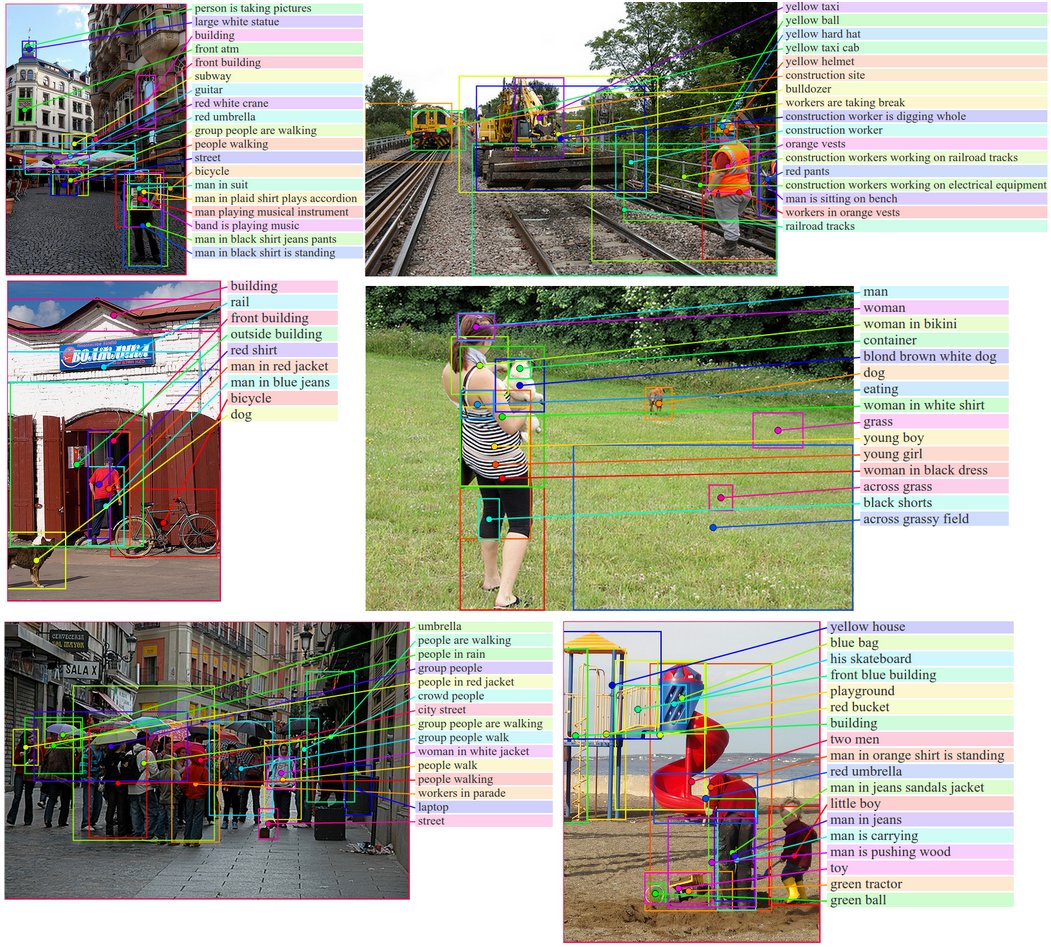}
\caption{Additional examples of region captions on the test set of Flickr30K.}
\label{fig:region2}
\end{figure*}

Additional examples of region annotations can be found in Figure \ref{fig:region2}. Note that we annotate regions based on the content of each image region alone, which can cause erroneous predictions when not enough context is available in the bounding box (e.g. a generated description that says ``container'' detected on the back of a dog's head in the image on the right, in the second row). We found that one effective way of using the contextual information and improving the predictions is to concatenate the fullframe feature CNN vector to the vector of the region of interest, giving 8192-dimensional input vector the to RNN. However, we chose to omit these experiments in our paper to preserve the simplicity of the mode, and because we believe that cleaner and more principled approaches to this challenge can be developed.

\subsection{Training the Multimodal RNN}

There are a few tricks needed to get the Multimodal RNN to train efficiently. We found that \textbf{clipping the gradients} (we only experimented with simple per-element clipping) at an appropriate value consistently gave better results and helped on the validation data. As mentioned in our paper, we experimented with SGD, SGD+Momentum, Adadelta, Adagrad, but found \textbf{RMSProp} to give best results. However, some SGD checkpoints usually also converged to nearby validation performance vicinity. Moreover, the distribution of the words in English language are highly non-uniform. Therefore, the model spends the first few iterations mostly learning the biases for the Softmax classifier such that it is predicting every word at random with the appropriate dataset frequency. We found that we could obtain faster convergence early in the training (and nicer loss curves) by explicitly \textbf{initializing the biases} of all words in the dictionary (in the Softmax classifier) to log probability of their occurrence in the training data. Therefore, with small weights and biases set appropriately the model right away predicts word at random according to their chance distribution. After submission of our original paper we performed additional experiments with comparing an RNN to an LSTM and found that \textbf{LSTMs} consistently produced better results, but took longer to train. Lastly, we initially used word2vec vectors as our word representations $x_i$, but found that it was sufficient to train these vectors from random initialization without changes in the final performance. Moreover, we found that the word2vec vectors have some unappealing properties when used in multimodal language-visual tasks. For example, all colors (e.g. red, blue, green) are clustered nearby in the word2vec representation because they are relatively interchangeable in most language contexts. However, their visual instantiations are very different.

\clearpage

\begin{table*}[t]
\small
\centering
\begin{tabulary}{\linewidth}{L|CCCC|CCCC}
& \multicolumn{4}{c}{Image Annotation} & \multicolumn{4}{c}{Image Search} \\
\textbf{Model} & \textbf{R@1} & \textbf{R@5} & \textbf{R@10} & \textbf{Med} \it{r} & \textbf{R@1} & \textbf{R@5} & \textbf{R@10} & \textbf{Med} \it{r} \\
\hline
\multicolumn{9}{c}{\textbf{Flickr8K}} \\
\hline
DeViSE (Frome et al. \cite{frome2013devise}) & 4.5 & 18.1 & 29.2 & 26 & 6.7 & 21.9 & 32.7 & 25 \\
SDT-RNN (Socher et al. \cite{sochergrounded}) & 9.6 & 29.8 & 41.1 & 16 & 8.9 & 29.8 & 41.1 & 16 \\
Kiros et al. \cite{kiros2014unifying} & 13.5 & 36.2 & 45.7 & 13 & 10.4 & 31.0 & 43.7 & 14 \\
Mao et al. \cite{mao2014explain} & 14.5 & 37.2 & 48.5 & 11 & 11.5 & 31.0 & 42.4 & 15\\
DeFrag (Karpathy et al. \cite{defrag}) & 12.6 & 32.9 & 44.0 & 14 & 9.7 & 29.6 & 42.5 & 15 \\
Our implementation of DeFrag \cite{defrag} & 13.8 & 35.8 & 48.2 & 10.4 & 9.5 & 28.2 & 40.3 & 15.6\\
Our model: DepTree edges & 14.8 & 37.9 & 50.0 & 9.4 & 11.6 & 31.4 & 43.8 & 13.2 \\
Our model: BRNN & \textbf{16.5} & \textbf{40.6} & \textbf{54.2} & \textbf{7.6} & \textbf{11.8} & \textbf{32.1} & \textbf{44.7} & \textbf{12.4} \\
\end{tabulary}
\vspace{0.05in}
\caption{{Ranking experiment results for the Flickr8K dataset.}}
\label{tab:8k}
\vspace{-0.1in}
\end{table*}

\end{document}